\documentclass{article}

\usepackage{PRIMEarxiv}
\usepackage{amsmath,amssymb,amsfonts} 
\usepackage[utf8]{inputenc} 
\usepackage[T1]{fontenc}    
\usepackage{hyperref}       
\usepackage{url}            
\usepackage{booktabs}       
\usepackage{amsmath,amssymb,amsfonts}      
\usepackage{nicefrac}       
\usepackage{microtype}      
\usepackage{lipsum}
\usepackage{fancyhdr}       
\usepackage{graphicx}       
\graphicspath{{media/}}     
\usepackage{algorithmicx}
\usepackage{algorithm}
\usepackage{caption}
\usepackage{graphicx}
\usepackage{textcomp}
\usepackage{algpseudocode}
\usepackage{bm}
\usepackage{float}
\usepackage{algpseudocode}
\usepackage{multirow}
\usepackage{subcaption}
\usepackage{cleveref} 
\usepackage{color}
\graphicspath{{./fig/}}

\algnewcommand{\LineComment}[1]{\State \(\triangleright\) #1}
\newcommand{\RNum}[1]{\uppercase\expandafter{\romannumeral #1\relax}}

\def\BibTeX{{\rm B\kern-.05em{\sc i\kern-.025em b}\kern-.08em
    T\kern-.1667em\lower.7ex\hbox{E}\kern-.125emX}}

\author{
WANG Ding
}

\title{Gate Recurrent Unit for Efficient Industrial Gas Identification}
\begin{document}

\maketitle

\begin{abstract}
\textcolor{red}{This is the first draft, the final version will submit to the IEEE for possible publication. Copyright may be transferred without notice, after which this version may no longer be accessible.} In recent years, gas recognition technology has received considerable attention. Nevertheless, the gas recognition area has faced obstacles in implementing deep learning-based recognition solutions due to the absence of standardized protocols. To tackle this problem, we suggest a new GRU. Compared to other models, GRU obtains an higher identification accuracy.
\end{abstract}


\section{Introduction}
Gas sensing technology has undergone substantial advancements in the last century, transitioning from the reliance on canaries to utilizing contemporary MEMS and optical gas sensors. This field has evolved to cater to the requirements of diverse applications, such as industrial\cite{b1}, medical\cite{b4}, food\cite{b5}, and environmental sectors\cite{b6}. Gas recognition technology can be categorized into two main tasks: identifying the components of a gas and estimating its concentration. gas categorization tasks always offer broader application possibilities than gas concentration estimation jobs.

Due to the growing complexity of gas sensing applications, researchers have dedicated significant attention to enhancing gas mixture identification algorithms.
Pan and colleagues introduced a compact multiscale convolutional neural network with attention to tackle problems such as inadequate selectivity and drift\cite{b11}. Yang and colleagues devised a temporal convolution network fitted with a transformer to detect ternary combinations\cite{b12}. 

Although these strategies have been highly successful in controlled environments, they face extra difficulties when applied in real-world scenarios. Various algorithms' efficacy continues to be difficult because of the diverse research priorities and datasets.

Based on this analysis, we developed an algorithm designed to optimize sensor usage while maintaining high recognition accuracy: the GRU . GRU achieved recognition accuracy of 98.33\% for two-sensor data.

The remainder of this article is structured as follows. Section \uppercase\expandafter{\romannumeral2} presents the proposed methodology, Section \uppercase\expandafter{\romannumeral3}  discusses experimental verification and results, and Section  \uppercase\expandafter{\romannumeral4} concludes the paper.

\section{Methodology}

\subsection{Static Sensor Array System Design}
The constructed machine olfaction system primarily consists of an eight-sensor gas array, a +5V DC-regulated power supply, and a LabVIEW data acquisition interface. This study utilizes the built-in 8-channel 12-bit A/D converter of the STM32 microcontroller to sample the dynamic signals from the sensor array in real time. The gas sensor array includes eight different metal oxide semiconductor (MOS) gas sensors: MQ135, TGS813, TGS2611, TGS2610, TGS2620, TGS2600, TGS2602, and MP503. The output signals of these sensors correspond to channels 0-7 of the A/D converter, respectively.

\subsubsection{Data Acquisition Process}
We have turned off six sensors from the present eight-sensor array and have chosen to keep two sensors, namely TGS813 and TGS2611, for data collecting. We collected data for hydrogen in its pure form, carbon monoxide in its pure form, and a combination of hydrogen and carbon monoxide. The concentration of each gas was evenly dispersed throughout the range of 10 to 1000 parts per million (ppm). A total of 150 data points were obtained for each gas. The gas mixture's concentration ranged from 10 to 1000 parts per million (ppm). The hydrogen concentration in the gas mixture exhibited a random distribution, whereas the carbon monoxide concentration demonstrated a normal distribution of the hydrogen concentration. The collection period for each sample was 200 seconds.

\subsection{ GRU}

GRU comprises three modules: the external attention module, the GRU module, and the decoder.

\subsubsection{External Attention Mechanism}
This work incorporates external attention techniques to augment the model's concentration on pertinent features. In contrast to traditional self-attention, external attention incorporates a supplementary memory module to store and retrieve crucial information. This enhances both attention span and accuracy. This enables the model to prioritize important input features and maintain performance when dealing with complicated, high-dimensional data.

\subsubsection{GRU Mechanism}
The Gated Recurrent Unit (GRU) \cite{b26}, a classic model for time series data analysis, has already been successfully applied in several gas recognition algorithms \cite{b27}. GRU is a recurrent neural network (RNN) architecture designed to address the vanishing gradient problem inherent in traditional RNNs. GRUs use gating mechanisms, including an update gate and a reset gate, to efficiently capture long-term dependencies in sequential data. 

\subsection{Evaluation Metrics}

We evaluated the model using conventional methods such as precision, recall, and F1 score.

The weighted average is a crucial method for evaluating multi-class models, particularly suitable for situations where class distributions are imbalanced. The weighted average calculates the overall precision, recall, and F1-score by considering the number of samples for each class, also called support, thus providing a more accurate reflection of the model's overall performance.

For each class, we first calculate its precision, recall, and F1-score, then weigh these metrics with the support of that class. Suppose there are $n$ classes, where the support for class $i$ is $s_i$, the precision is $P_i$, the recall is $R_i$, and the F1-score is $F_i$. In this paper, $n = 3$. 

This calculation ensures that each class's weight in the overall performance evaluation corresponds to its proportion in the dataset, thereby avoiding biases caused by classes with extremely large or small sample sizes.

\section{Experimental Results and Analysis}
\subsection{Description of Dataset}

The dataset includes three types of gases: pure hydrogen, pure carbon monoxide, and a mixture of hydrogen and carbon monoxide. Each pure gas has 150 samples, while the gas mixture has 300 samples, totaling 600. Each sample contains 2000 nodes corresponding to 200 seconds. 

We stratified the processed data based on the gas type at a ratio of 8:2 and randomly divided it into training-validation and test sets. To compensate for the small size of the dataset, we used five-fold cross-validation for the training-validation set. This means that for each epoch, we selected the model with the highest validation accuracy and tested the model on the test set.

In addition, we extracted data from the TGS2611 sensor to generate individual sensor data for comparative validation. The training process, including the hyperparameters, remained the same for the two-sensor data.

\subsection{Model Training Description}
The model uses a batch size of 24 to utilize data during training efficiently. Feature extraction involves a db5 wavelet transform with one layer to capture essential patterns in the time series data. The attention mechanism with 500 hidden units focuses on crucial parts of the input. A 3-layer GRU with eight hidden units and dropout processes the sequential data, while the decoder with GeLU activation and dropout generates the output. A distributed decay learning rate starting at 0.0005 and a cross-entropy loss function guide the training process.

\subsection{Results Analyzation}

\begin{table*}[!h]
\centering
\caption{Results of Models}
\label{results}
\resizebox{0.6\textwidth}{!}{
\begin{tabular}{ccccccccc}
\toprule
\multirow{2}{*}{Models}
& \multicolumn{4}{c}{Two Sensors} & \multicolumn{4}{c}{One Sensor} \\ \cline{2-9} 
& Accuracy & Precision  & Recall & F1  & Accuracy & Precision & RRecall &  F1   \\ \hline
GRU & 0.9833 & 0.992 & 0.983  & 0.0167 & 0.8917 & 0.893 & 0.891 &0.891  \\ \hline
SVM& 0.6583 & 0.7422 & 0.655  & 0.325 & 0.6333 & 0.641 & 0.633 & 0.554  \\ \hline
RF& 0.7917 & 0.7993 & 0.7935  & 0.2 & 0.7583 & 0.755 & 0.758 & 0.754  \\ \hline
KNN& 0.7917 & 0.8044 & 0.785  & 0.2 & 0.6333 & 0.635 & 0.618 & 0.617   \\ \bottomrule
\end{tabular}
}
\end{table*}

We compared the results from several models using data from two sensors and one sensor separately. The results can be found in Table \ref{results}. This comparison included three traditional machine learning models: Support Vector Machines (SVM), Random Forests (RF), and K-nearest neighbors (KNN).

Table \ref{results} presents the test results of our model compared to other models, including accuracy, precision, recall, F1 score, cross-recognition rate, and confusion rate. Our algorithm achieved a remarkably high recognition accuracy of 98.33\% when classifying using data from two sensors, while conventional machine learning models exhibited a noticeable performance gap.

Regarding mixed recognition using data from a single sensor, our algorithm demonstrated superior performance compared to traditional machine learning models. They showed a smaller decrease in accuracy when using single-sensor data.

These results highlight the efficacy of our proposed GRU model in accurately recognizing gas mixtures with high reliability. The superior performance of our model, especially in challenging single-sensor scenarios, underscores its robustness and potential for real-world applications.

Based on the above results, we believe that the core component of a gas sensor capable of meeting the demands of natural industrial environments should not merely be neural network modules like RNNs. The previous brute-force approach is impractical in industrial settings that require low power consumption and compact size. We propose that a crucial future direction for gas recognition models is to extract information from sensor signals more effectively and achieve superior performance with fewer sensors.

\section{Conclusion}
This paper analyzed the requirements for gas mixture recognition algorithms in industrial environments. We introduced a novel model, the GRU, which achieved superior results compared to other models, with data from two sensors and a single sensor. These results suggest that the GRU model offers a promising approach for enhancing gas recognition in practical industrial applications.

\end{document}